\documentclass[conference]{IEEEtran}
\IEEEoverridecommandlockouts
\usepackage{cite}
\usepackage{amsmath,amssymb,amsfonts}
\usepackage{algorithmic}
\usepackage{graphicx}
\usepackage{textcomp}
\usepackage{xcolor}
\def\BibTeX{{\rm B\kern-.05em{\sc i\kern-.025em b}\kern-.08em
    T\kern-.1667em\lower.7ex\hbox{E}\kern-.125emX}}

\usepackage[]{graphicx}
\usepackage{url}
\usepackage{hyperref}

\usepackage{amsmath}
\DeclareMathOperator*{\argminA}{arg\,min}

\begin{document}

\title{Multitask Learning \\ for Network Traffic Classification
}

\author{\IEEEauthorblockN{Shahbaz Rezaei}
\IEEEauthorblockA{\textit{Department of Computer Science} \\
\textit{University of California} \\ Davis, USA\\
Email: srezaei@ucdavis.edu}
\and
\IEEEauthorblockN{Xin Liu}
\IEEEauthorblockA{\textit{Department of Computer Science} \\
\textit{University of California} \\ Davis, USA\\
Email: liu@cs.ucdavis.edu}}

\maketitle

\begin{abstract}
Traffic classification has various applications in today's Internet, from resource allocation, billing and QoS purposes in ISPs to firewall and malware detection in clients. Classical machine learning algorithms and deep learning models have been widely used to solve the traffic classification task. However, training such models requires a large amount of labeled data. Labeling data is often the most difficult and time-consuming process in building a classifier. To solve this challenge, we reformulate the traffic classification into a multi-task learning framework where bandwidth requirement and duration of a flow are predicted along with the traffic class. The motivation of this approach is twofold: First, the bandwidth requirement and duration are useful in many applications, including routing, resource allocation, and QoS provisioning. Second, these two values can be obtained from each flow easily without the need for human labeling or capturing flows in a controlled and isolated environment. We show that with a large amount of easily obtainable data samples for bandwidth and duration prediction tasks, and only a few data samples for the traffic classification task, one can achieve high accuracy. Therefore, our proposed multi-task learning framework obviates the need for a large labeled traffic dataset.
We conduct two experiments with ISCX and QUIC public datasets and show the efficacy of our approach.
\end{abstract}


%

\section{Introduction}
\label{sec:intro}

Network traffic classification has a wide variety of applications in today's Internet, such as resource allocation, QoS provisioning, billing in ISPs, anomaly detection, etc. The earliest approaches to solve network traffic classification used port numbers or unencrypted packet payloads. These methods relied on human labor for continuously finding patterns in unencrypted payloads or matching port numbers. Due to inefficiency and lack of accuracy, new methods based on classical machine learning algorithms emerged, such as such as random forest (RF) and k-nearest neighbor (KNN).

For several years, classical machine learning algorithms had achieved state-of-the-art accuracy in the traffic classification task. However, these relatively simple methods were not able to capture more complex patterns which exist in today's Internet traffic and, therefore, their accuracy has degraded. Recently, deep learning models achieved the state-of-the-art performance in traffic classification. Their ability to learn complex patterns and perform automatic feature extraction makes them desirable for traffic classification. 

Although deep learning methods can achieve high accuracy, they require a large amount of labeled training data. In the network traffic classification task, labeling is a time-consuming and cumbersome task. In order to correctly label each flow, researchers usually capture flows of each class in isolation and in a controlled environment with minimum background traffic. This process is time-consuming and labor-intensive. Moreover, traffic patterns observed in a controlled environment might differ significantly from real traffic, which makes the inference inaccurate.

To mitigate the need for a large amount of labeled training samples, we propose a multi-task learning approach which performs three predictions (tasks), for which only one requires human effort and controlled environment for labeling. These are bandwidth, duration, and traffic class prediction tasks. For any captured data, whether it is captured in a controlled environment in isolation or not, one can easily compute the total bandwidth and duration of each flow without human labeling. Hence, by formulating the traffic classification problem in a multi-task learning framework where the large amount of model parameters are shared among all tasks, one can train the model with a large amount of data for bandwidth and duration tasks and only a small number of labeled samples for traffic class prediction task. Moreover, for various applications, such as resource allocation or QoS purpose, bandwidth and duration prediction is highly useful.

\section{Related Work}

Before the widespread emergence of deep learning models, classical machine learning approaches have been widely used for network traffic classification \cite{velan2015survey}. These methods usually relied on supervised learning methods, such as support vector machine (SVM) \cite{kumano2014towards,khakpour2013information}, C4.5 \cite{alshammari2011can,alshammari2010investigation}, naive Bayes \cite{okada2011application,sun2010novel}, k-nearest neighbor \cite{bar2010realtime,wright2006inferring}, etc., or unsupervised clustering methods, such as k-means \cite{zhang2012encrypted,du2013design} or Gaussian mixture model \cite{bernaille2007early}. However, their accuracy has declined recently due to the simplicity, manual feature extraction (which becomes more difficult with today's strongly encrypted traffic), and the lack of high learning capacity to capture more complex patterns.

In the past few years, with the promising success of deep learning methods on variety of problems, such as image classification, voice recognition, translation, etc., network researchers recently adopted these methods for traffic classification \cite{wang2019survey}. In \cite{zhou2017method}, a LeNet-5 convolutional Neural Network (CNN) model, designed in 1998 for handwritten numeral recognition, is used for traffic type classification. Numerous statistical features are rearranged into a 2-dimensional image as input to the model. They report high accuracy, but the model cannot be used for online applications \cite{rezaei2019deep} because it requires an entire flow to be observed to obtain the statistical features. In \cite{tong2018novel}, authors use both statistical features and payload data for traffic classification of QUIC protocol. They first use statistical features with random forest algorithm to distinguish between chat and voice call with other classes. If other classes are detected, they use payload data with a CNN model to classify video streaming, file transfer, and Google music. Their first stage needs the entire flow to be observed and, hence, it is only suitable for offline applications. Payload information, although encrypted, has been used in other papers as well. In \cite{lotfollahi2017deep}, a CNN and stacked Auto-Encoder (SAE) are used together on ISCX dataset to classify traffic types and applications. These methods use deep neural networks as a black box without identifying human understandable features.

In \cite{chen2017seq2img}, time-series features of each flow are converted into 2-dimensional images using Reproducing Kernel Hilbert Space (RKHS). The produced images are used as input to a CNN model. They compare their CNN model with classical machine learning approaches, including SVM, decision tree and naive Bayes. The CNN model achieve over $99\%$ accuracy and outperforms classical machine learning approaches. In \cite{lopez2017network}, a convolutional neural network, a long short-term memory (LSTM) model and various combinations are used for classification of several services, such as YouTube and Office365. They achieve the accuracy of around $96\%$ when time-series features and header features are used with the CNN/LSTM architecture. In \cite{rezaei2019large}, CNN and CNN+LSTM models are used for identification of 80 mobile applications. They use raw packet header and payload and achieve high accuracy for large number of classes. They used occlusion analysis to elucidate how deep learning models can classify encrypted traffic for the first time. They show that unencrypted handshake fields in SSL/TLS protocol can be effectively used for app identification. However, they use large dataset captured at an operational ISP to train such a large-scale model.

In \cite{rezaei2019deep}, a general framework is proposed providing a straightforward guidelines and directions for any traffic classification task. Most previous work falls under the general framework. However, all these methods rely on supervised learning and require a large amount of labeled data for training. This becomes more problematic for deep models because they need significantly more training data than classical machine learning approaches.

The only study that addresses the need for a large labeled dataset is \cite{rezaei2019achieve}. This approach consists of a semi-supervised learning method, where a CNN model is first pre-train to predict several statistical features from the sampled packets. They use time-series features of sampled packets. Then, they replace the last few layers with new ones and then re-train with a small labeled dataset. The advantage of their approach is that it does not need human effort for labeling the pre-trained dataset because statistical features can be computed easily when entire flows are available. However, their approach takes sampled data packets which means that it needs to observe a large portion of a flow before performing the classification which is not suitable for online applications. In this paper, we propose a multi-task learning approach that outperforms both single-task learning and transfer learning.

\section{Deep Learning Background}
\subsection{Convolutional Neural Networks (CNNs)}

\begin{figure}
	\centering
	\includegraphics[width=3in]{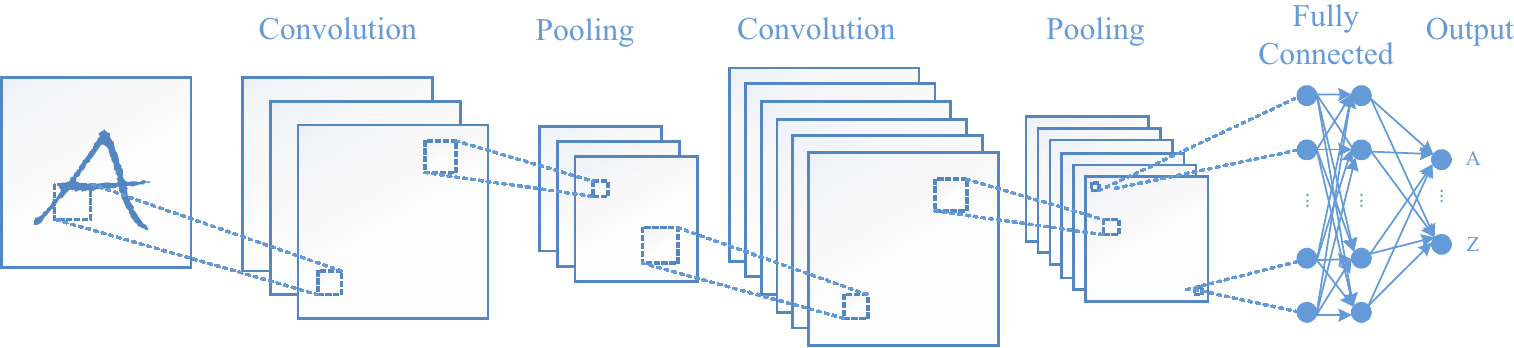}
	\caption{A typical architecture of CNN models \cite{rezaei2019achieve} }
	\label{fig-CNN}
\end{figure}

Convolutional neural networks (CNNs) are types of deep learning models consisting of several layers that use convolution operations. The architecture of CNN models is inspired by the organization of animal visual cortex. CNNs consist of several convolution layers, pooling layers, and often fully connected (FC) layers, as shown in Fig.~\ref{fig-CNN}. In a convolution layer, a set of small kernels with a small number of learnable parameters are used to capture patterns from the output of the previous layer. To generate an output, a convolution layer uses the same set of kernels on the entire input.  By using the same set of kernels in a layer, the number of learnable parameters are dramatically reduced. Moreover, the use of these kernels on the entire input helps the model to also capture shift invariant features more easily. For example, in an image classification task, a kernel that captures the pattern of a tiger skin can detect the pattern regardless of the location of the tiger in an image. This is particularly helpful for tasks that are inherently shift invariant, including network traffic classification, where some patterns may occur at the first few packets, at the end of the flow, or potentially at any part of the flow. Another layer that is usually used in CNN models is the pooling layer which is mainly responsible for subsampling. At the end of the set of convolution and pooling layers, a set of fully connected layers are often used to capture high-level features of an input.

\subsection{Multi-task Learning (MTL)}
Multi-task learning (MTL) aims to perform several learning tasks simultaneously under the assumption that the tasks are not completely independent and one can improve the learning of another. For instance, detecting dangerous objects and a distance-based danger assessment are two tasks important to autonomous driving. Since these two tasks are related and can benefit from a shared representation, one can define a multi-task learning approach to jointly learn these tasks \cite{chen2018multi}.

The most common approach to multi-task learning is hard parameter sharing where some parameters of the deep learning models are shared among tasks and some parameters are kept task-specific \cite{rezaei2019security}. Our hard parameter sharing model is explained in Section~\ref{model-arc}. The multi-task learning model is more effective than several single-task learning models because in MTL datasets of all tasks can help all other tasks to be learned better. In this paper, we will show that we can improve the accuracy of the traffic classification task by using a multi-task learning approach where the data are easy to obtain for other tasks, namely predicting bandwidth and duration of a flow. Hence, we show that the accuracy of traffic classification task improves significantly when abundant data of other tasks is used when training a MTL framework.

\section{Methodology}

\subsection{Motivation}

This paper is motivated by the following observations: First, it is difficult to obtain sufficient labeled training data for traffic classification task. At the same time, there are other prediction tasks that are needed for resource allocation, have easy-to-obtain labels, and can be used to improve the accuracy of traffic classification task. Therefore, we are motivated to solve the two problems together through multi-task learning.

First, capturing a large enough labeled dataset for traffic classification to train a deep model is a time-consuming and cumbersome task \cite{rezaei2019deep}. Moreover, correctly labeling captured data is also challenging, particularly when background traffic or more than one traffic classes exist during capturing. On the other hand, unlabeled data is often abundant and easy to capture. Hence, it is desirable to be able to use a large amount of unlabeled data to dramatically reduce the number of labeled data needed for training. In this paper, we use a large unlabeled datasets with a small number of labeled data in a multi-task learning framework to solve this problem.

The second motivation of our approach is that ISPs or data centers often perform traffic classification for billing, resource allocation or QoS purposes. For such purposes, the traffic class along with other flow features, such as bandwidth requirement and duration, can significantly improve ISPs decision for resource allocation, QoS, etc. In this paper, we propose a multi-task learning approach that takes the first few time-series features of a flow and perform three prediction tasks: bandwidth, duration, and traffic class. Not only the bandwidth and duration predictions are useful, they do not need human labor for labeling and one can capture a large amount of flows and then compute the bandwidth and duration of them easily.

\subsection{Datasets}
\label{sec-datasets}

\subsubsection{QUIC Dataset}

The QUIC dataset \cite{rezaei2019achieve} is captured at University of California at Davis. It contains QUIC traffic of 5 Google services: Google Doc (1251 flows), Google Drive (1664 flows), Google Music (622 flows), Youtube (1107 flows), Google Search (1945 flows). The dataset contains time-series features: packet length, relative time, and direction. The dataset has already been pre-processed. According to \cite{rezaei2019achieve}, all short flows that have fewer than 100 packets had been removed. Note that all flows in the dataset are labeled. However, to evaluate our multi-task learning approach, we only use a small portion of class labels during training.

\subsubsection{ISCX VPN-nonVPN Dataset}

ISCX Dataset \cite{draper2016characterization} is captured at University of New Brunswick and it contains raw pcap files of several traffic types. The dataset provides fine-grained labels which allows different categorization: application-based (e.g. AIM chat, Gmail, Facebook, etc), traffic-type-based (e.g. chat, streaming, VoIP, etc), and VPN/non-VPN. In this paper, we divide the dataset into 5 categories that have different QoS requirements and bandwidth/duration characteristics: chat, email, file transfer, streaming, and VoIP. Both UDP and TCP traffic exist in the dataset. For TCP flows, we look for FIN packet to identify the end of TCP flows. For UDP flows, we use flow timeout of 15 seconds as in \cite{lashkari2017characterization} to mark the end of UDP flows. Similar to the QUIC dataset, all flows are associated with a traffic type label, but we only use a small portion of labels for traffic class prediction of multi-task learning.

\subsection{Input Features and Prediction Outputs}
\label{sec-inp-out}
In general, modern network traffic classifiers use one or a combination of four categories of input features: time-series, header, payload, and statistical features \cite{rezaei2019deep}. Header information is rarely used nowadays because it does not achieve high accuracy. Statistical features are obtained from the entire flow and, consequently, is not suitable for online classification where the prediction is needed as soon as traffic emerges. Online classification is necessary when resource allocation, QoS or routing decisions depend on the prediction output. Payload data has been shown to be useful for some datasets and special traffic types and encryption methods \cite{lotfollahi2017deep, aceto2018mobile}. The success of these methods stems from unencrypted fields during the handshake phase of TLS 1.2 \cite{rezaei2019achieve}. However, modern encryption methods, e.g. QUIC and TLS 1.3, reduce the number of unencrypted fields as much as possible. Hence, for the new and stronger encryption protocols, payload information itself may not be as useful. Note that many traffic classification approached used statistical features of the entire flow. Bandwidth requirement and duration of a flow can also be considered as statistical feature which can be obtained by observing the entire flow. However, in our proposed method, we predict the bandwidth and duration because we can only observe the first few packets, not the entire flow. Hence, we treat bandwidth and duration as separate tasks as output, in contrast to common traffic classification methods where they use these values as input.

In this paper, we use three time-series features, that is, packet length, inter-arrival time, and direction, of the first $k$ packets. The input of our model is a vector of length $k$ with 2 channels. The first channel contains the inter-arrival time of the first $k$ packets and the second channel contains the length and direction combined. As in \cite{rezaei2019achieve}, for the second channel, a positive value indicates the packet length in forward direction (from a client to a server) and a negative value indicates the packet length in backward direction. Moreover, we normalize the data by assuming a maximum value of 1434 Bytes for length and 1 second for inter-arrival time.

\begin{figure}
	\centering
	\includegraphics[width=3.0in]{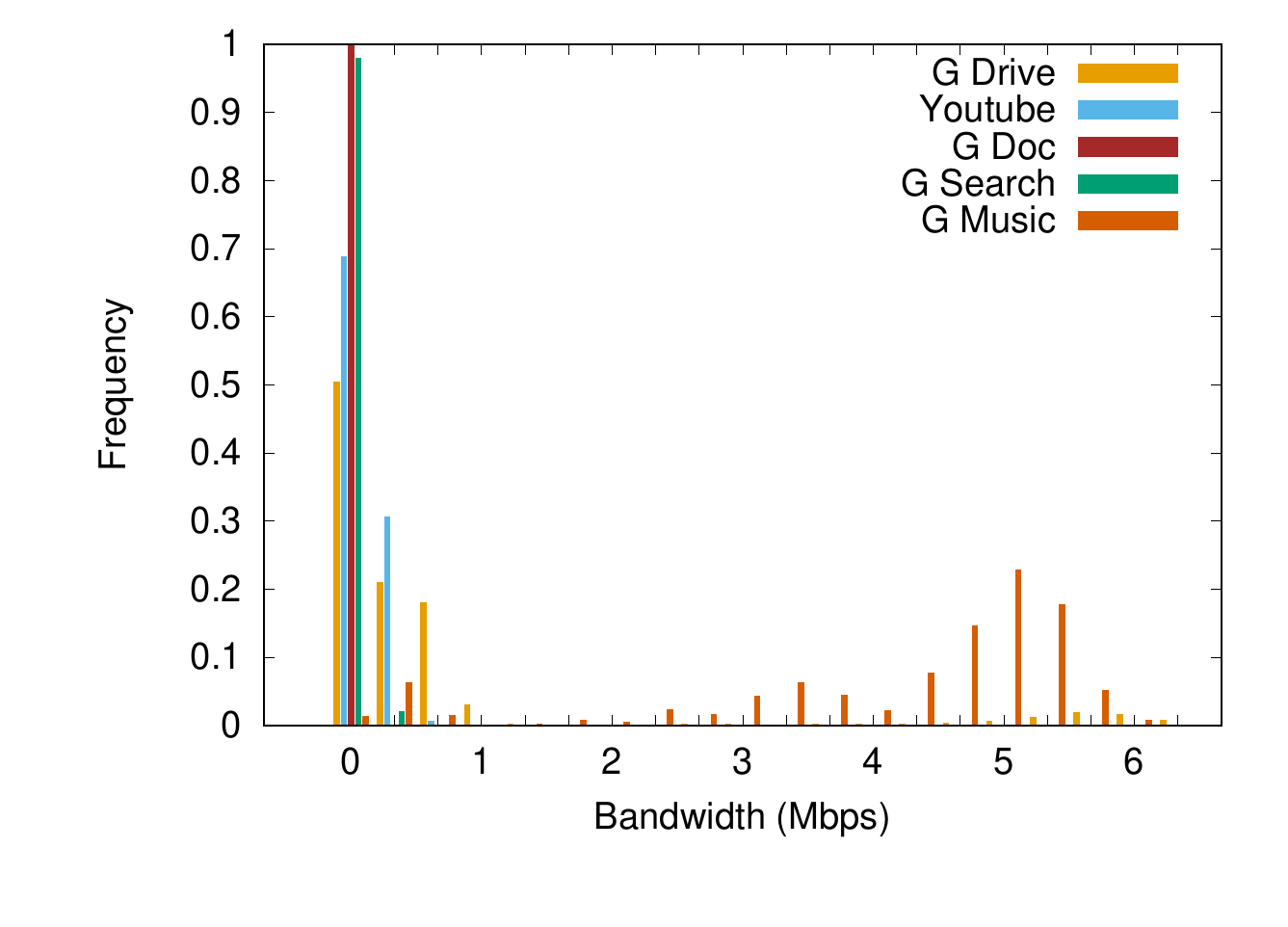}
	\caption{Histogram of Bandwidth}
	\label{fig-bw-quic}
\end{figure}

\begin{figure}
	\centering
	\includegraphics[width=3.0in]{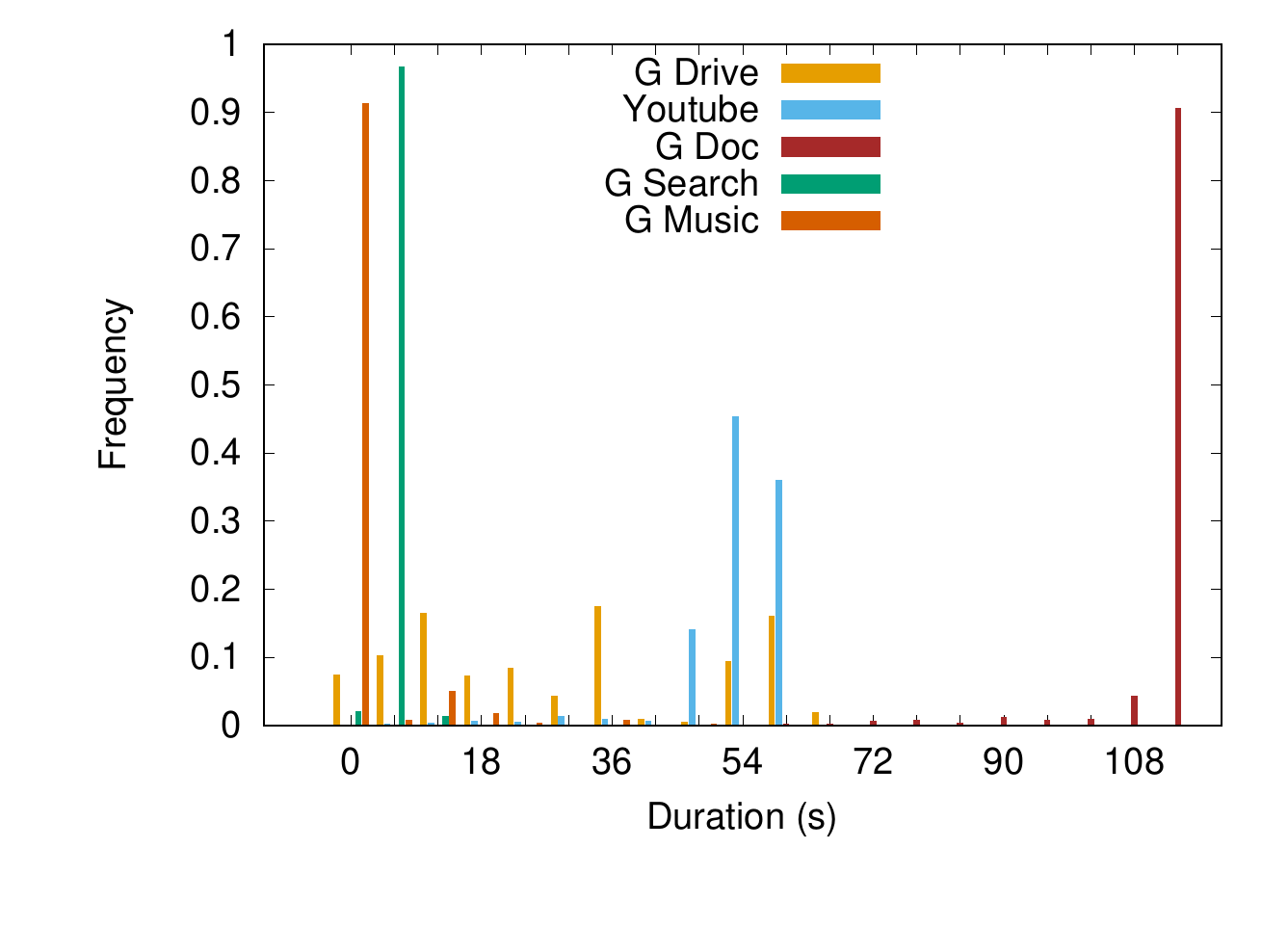}
	\caption{Histogram of Duration}
	\label{fig-duration-quic}
\end{figure}

The common approach for traffic classification often focuses on predicting traffic types, applications, operating systems, user actions, etc. In addition to such class labels, we aim to predict duration and bandwidth requirements of traffic which can be used for resource allocation, routing, or QoS purposes. In this paper, we formulate the bandwidth and duration prediction problems as classification instead of regression. In our experiment, the model takes a long time to converge and does not perform well when bandwidth and duration tasks are defined as regression problems. In addition, coarse-grained predictions are often enough for routing or QoS purposes. Therefore, we reformulate these tasks as classification tasks.

The labels of bandwidth and duration class definitions are shown in Table \ref{tbl-BD-classes}. We divide the bandwidth and duration values into five classes. We call $[bw_1, ..., bw_4]$ and $[d_1, ..., d_4]$ bandwidth and duration divider. For example, if the bandwidth of a flow is between $bw_1$ and $bw_2$, the class number 2 is assigned as a label to that flow. The number of classes for bandwidth and duration prediction tasks can be different from the number of traffic classes. They may depend on the application, scenario and ISP's needs. For example, an ISP that only considers the difference between short-lived and long-lived flows may only define two duration classes. We experimentally show in the evaluation section that if the classes are defined such that the bandwidth and duration classes correspond to the average values of each traffic class, the accuracy of all tasks improves. However, we show that even choosing an arbitrary vales for categorization of bandwidth and duration in the proposed multi-task framework outperforms a single-task classifier.

\begin{table}[t]
	\centering
	\caption{Bandwidth and duration classes definition}
	\label{tbl-BD-classes}
	\begin{tabular}{*{15}{c}}
		\hline
		Class number & Bandwidth (B) & Duration (D) \\
		\hline
		Class 1 & $B < bw_1$ & $D < d_1$ \\
		\hline
		Class 2 & $bw_1< B < bw_2$ & $d_1< D < d_2$ \\
		\hline
		Class 3 & $bw_2< B < bw_3$ & $d_2< D < d_3$ \\
		\hline
		Class 4 & $bw_3< B < bw_4$ & $d_3< D < d_4$ \\
		\hline
		Class 5 & $B > bw_4$ & $D > d_4$ \\
		\hline
	\end{tabular}
\end{table}

To better understand the distribution of bandwidth and duration, we illustrate the histogram of bandwidth and duration for QUIC dataset classes in Fig.~\ref{fig-bw-quic} and Fig.~\ref{fig-duration-quic}, respectively. For bandwidth, one class (G Music) considerably consumes more bandwidth and is mostly larger than 1 mbps, while other classes vary between 1kbps and 1mbps. The reason why G Music bandwidth is considerably larger is because it attempts to download the entire track when it is played. This is more obvious by observing the duration of G Music flows in Fig.~\ref{fig-duration-quic}, which shows the very short-lived nature of G Music flows. Duration of different classes are relatively distinctive, despite overlaps in some regions. Note that based on the dataset, application, and ISP's need, bandwidth and duration categorization may change from one scenario to another. 

To find the optimal value for $[d_1, ..., d_4]$, the duration divider, we first find the average duration of each class. Then, we sort the average values and then find the middle point between two consecutive average values as $[d_1, ..., d_4]$. For instance, in QUIC dataset, average duration for G Music, G Search, G Drive, YouTube, and G Doc are [2.77, 9.83, 32.08, 56.44, 114.10] seconds. Hence, the duration divider is [6.30, 20.96, 44.26, 85.27]. Similar approach is also used to obtain the bandwidth divider array. Note that these values are the optimal values obtained from the entire dataset. However, our assumption is that only a small number of labeled data is available. Hence, in our evaluation, we obtain the dividers based on the small number of labeled samples. Therefore, the divider values are slightly different from the optimal values. However, we show in the evaluation section that using these dividers still improves the performance considerably.

In this paper, we use bandwidth and duration as tasks that do not need human effort for labeling and can be obtained easily in large quantities. One can also use other statistical features for such tasks, such as average packet length, standard deviation of inter-arrival time, etc., which are also used in the transfer learning approach in \cite{rezaei2019achieve}. Such tasks can also improve model training. However, these prediction tasks do not have direct usage and, thus, should be considered as auxiliary tasks.

\begin{figure}
	\centering
	\includegraphics[width=2.2in]{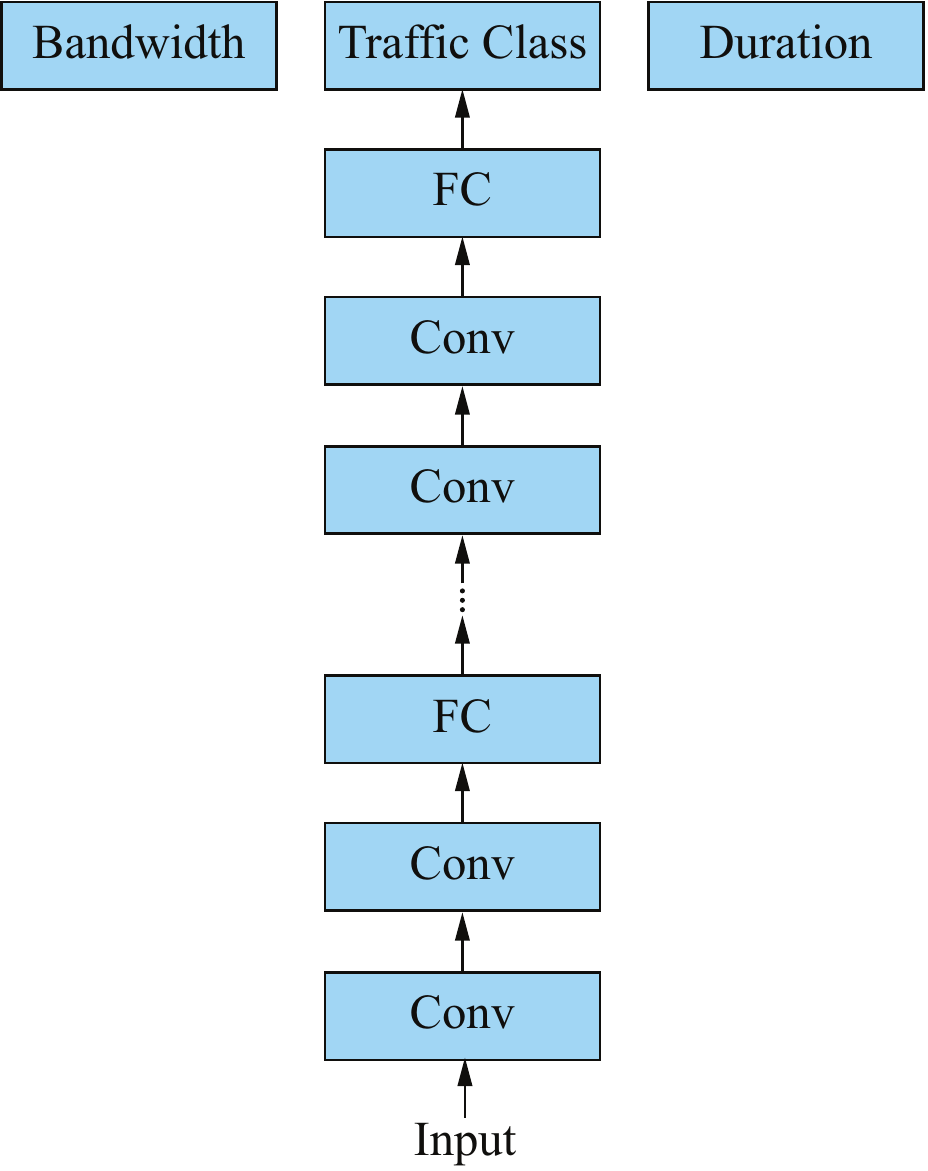}
	\caption{Multi-task learning model architecture}
	\label{fig-model}
\end{figure}

\subsection{Multi-task Model Architecture}
\label{model-arc}
In this paper, we use 1-dimensional convolutional neural network (CNN) in our multi-task learning model architecture. CNN architecture was first introduced and used for visual recognition tasks. However, in the past few years it has been extensively adopted for various tasks in other fields. One of the most important features of CNN models is their shift invariance, which is suitable for traffic classification task with time-series features because traffic patterns for each class may not necessarily appear in the same location in flows.

\begin{table*}[t]
	\centering
	\caption{Structure of the CNN model}
	\label{tbl-model-specification}
	\begin{tabular}{*{15}{c}}
		\hline
		- & Conv & Conv & Pool & Conv & Conv & Pool & Conv & Conv & Pool & FC & FC \\
		\hline
		Number of filters/neurons & 32 & 32 & - & 64 & 64 & - & 128 & 128 & - & 256 & 256 \\
		\hline
		Kernel size & 3 & 3 & 2 & 3 & 3 & 2 & 3 & 3 & 2 & - & - \\
		\hline
	\end{tabular}
\end{table*}

The overall architecture of our approach is shown in Fig. \ref{fig-model}. The details of the model parameters are presented in Table \ref{tbl-model-specification}. We use max pooling as it is commonly preferred over other pooling methods. Rectified linear unit (ReLU) activation is also used as an activation function in the entire model, except the last layers which contain Softmax.

Suppose bandwidth, duration and traffic class prediction tasks are denoted by B, D, and T, respectively. Additionally, we have $N$ training data for which $x_i$ represents the input of $i$-th data sample and $y_i^B$, $y_i^D$, and $y_i^T$ represent the corresponding output for bandwidth, duration, and traffic class prediction tasks\footnote{ In this paper, scalar, vector, and matrix are denoted by lowercase, bold lowercase, and bold capital letter, respectively.}. The objective of the multi-task learning approach can be formulated as 

\begin{multline*}
    \argminA_{\mathbf{W}^B, \mathbf{W}^D, \mathbf{W}^T} \sum_{i=1}^{N} \Big [ \ell (y_i^B, f(\boldsymbol{\mathbf{x}}_i;\mathbf{W}^B)) + \\
    \ell (y_i^D, f(\boldsymbol{\mathbf{x}}_i;\mathbf{W}^D)) + \lambda \ell (y_i^T, f(\boldsymbol{\mathbf{x}}_i;\mathbf{W}^T)) \Big ]
\end{multline*}
where $\ell$ is a cross entropy loss function. $\lambda$ is a weight that signifies the importance of the traffic class prediction task. Since this task has considerably fewer training data samples than the other two tasks, we can increase $\lambda$ to slightly compensate for the lack of labeled data. Note that for all training data, bandwidth and duration labels are available. However, only a small portion of data samples have traffic class labels. During training, we multiply the input of traffic class softmax layer to a mask vector to prevent back-propagation from this task for data samples that do not have a traffic class label.

\section{Evaluation}

\begin{table*}[t]
	\centering
	\caption{Accuracy on QUIC dataset}
	\label{tbl-QUIC-accuracy}
	\begin{tabular}{*{15}{c}}
		\hline
		\# of labeled samples & \multicolumn{3}{c}{Accuracy [Bandwidth, Duration, Traffic Class]}\\
		\hline
		 (For traffic class) & RF \cite{aceto2018mobile}  & CNN+RNN-2 \cite{lopez2017network} & Transfer learning & Multi-task learning\\
		\hline
		10 & [-, -, 48.67\%] & [89.33\%, 92.00\%, 64.67\%] & [-, -, 85.33\%] & [89.33\%, 91.00\%, 93.33\%]  \\
		20 & [-, -, 64.00\%] & [89.33\%, 92.00\%, 66.67\%] & [-, -, 87.33\%] & [90.33\%, 91.33\%, 94.67\%]  \\
		50 & [-, -, 78.00\%] & [89.33\%, 92.00\%, 76.67\%] & [-, -, 90.67\%] & [90.67\%, 91.33\%, 96.00\%]  \\
		100 & [-, -, 86.67\%] & [89.33\%, 92.00\%, 85.33\%] & [-, -, 92.67\%] & [90.67\%, 92.00\%, 97.33\%]  \\
		\hline
	\end{tabular}
\end{table*}

\subsection{Implementation Detail}
We use python and Keras package to implement our multi-task learning approach\footnote{ Codes are available at \href{https://github.com/shrezaei/MultitaskTrafficClassification}{https://github.com/shrezaei/MultitaskTrafficClassification}}. We use a server with Nvidia Titan Xp GPU and Intel Xeon W-2155 with Ubuntu 16.04. We conduct experiments with both ISCX and QUIC datasets. In all of our experiments, the training phase took less than a few minutes. We use batch optimization and Adam optimizer for training. The loss function and model architecture are explained in Section \ref{model-arc}.

\subsection{QUIC Dataset}
\label{sec-QUIC-eval}
In this section, we compare the accuracy of our multi-task learning approach with transfer learning and single-task learning approaches. Table \ref{tbl-QUIC-accuracy} shows the accuracy of all approaches for QUIC dataset. For single-task learning approach, we use two successful models, namely random forest (RF) with statistical features \cite{aceto2018mobile}, and CNN+RNN model proposed in \cite{lopez2017network}.
We train the model three times from scratch for each task.
For bandwidth and duration prediction, we use the entire dataset for training since it does not require human effort for labeling. That is why the accuracy remains the same when labeled samples are increased in Table \ref{tbl-QUIC-accuracy}. Note that the RF approach in \cite{aceto2018mobile} takes statistical features of the entire flow as input. Since the bandwidth and duration is statistical features, they can be obtained if the entire flow is accessible. So, there is no need for prediction of bandwidth and duration if the classifier can observe the entire flow. That is the reason we do not train RF models on these two tasks. For transfer learning, we deploy an approach similar to \cite{rezaei2019achieve} with slightly different source tasks. We first train the model with the entire dataset to predict the bandwidth/duration tuple. Note that for this task, we use both labels of the entire training set. After training the model, we remove the last layer and replace it with a new layer and initialized the weight, similar to \cite{rezaei2019achieve}. Then, we re-train the model for the traffic class prediction task. The final model only predicts the traffic class. That is the reason why Table \ref{tbl-QUIC-accuracy} does not contain the bandwidth and duration accuracy for the transfer learning approach. Note that there is another difference between the transfer learning approach we use and the one proposed in \cite{rezaei2019achieve}. In \cite{rezaei2019achieve}, the model takes \textit{sampled} time-series features as input which is not suitable for online classification needed for resource allocation, routing, or QoS purposes. Therefore, we take the first $k$ packets as input of the model without sampling. For multi-task learning approach, we train the entire model with all training data. We use the bandwidth and duration labels of the entire training data samples, while we only provide a limited number of labels for the traffic class task (specified in the first column). For this experiment, we set $\lambda$ to one to emphasize on all three tasks equally. Moreover, we use the first 60 packets as input ($k=60$).

As it is shown in Table \ref{tbl-QUIC-accuracy}, the accuracy of the traffic class prediction, for which we have limited labeled samples, is considerably higher with our multi-task learning approach than the transfer learning and single-task learning. In fact, the large amount of data that is available for bandwidth and duration tasks significantly improves the training process by allowing the model parameters to be trained with such abundant data. Although the transfer learning approach also reaps the benefits of the large dataset during pre-training, it is more prone to catastrophic forgetting \cite{wen2018overcoming}, losing the ability to perform previous tasks, or over-fitting, losing the ability to generalize by fitting closely to a training dataset, particularly when the target task has a small number of training samples.

In our experiments, we find that the accuracy of a single-task learning is $97.67\%$ when the entire dataset with all class labels are used. For traffic class prediction task, the multi-task learning approach with only 100 labeled samples reach almost the same accuracy as single-task learning using the entire labeled dataset. Hence, the multi-task learning approach can greatly reduce the number of labeled data. There is no significant performance difference between single-task learning and multi-task learning for bandwidth and duration prediction tasks because there are abundant data samples for these tasks.

\begin{figure}
	\centering
	\includegraphics[width=3.0in]{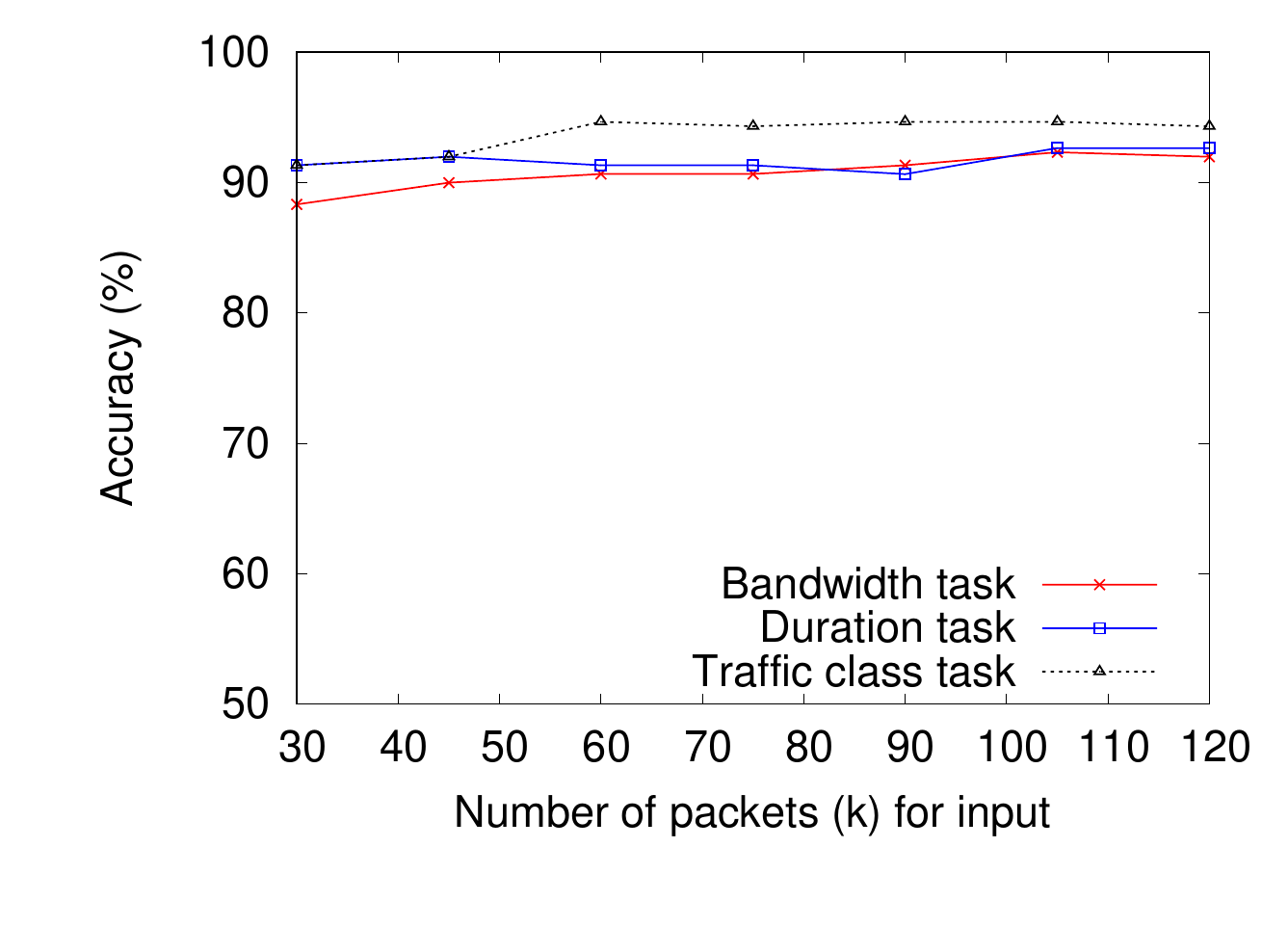}
	\caption{Number of input packets versus accuracy for QUIC dataset}
	\label{fig-acc-QUIC}
\end{figure}

Fig. \ref{fig-acc-QUIC} shows the accuracy of the multi-task learning approach for all three tasks when different number of packets are used as input\footnote{ Note that for the experiments with 30 and 45 packets, we removed one of the convolutional layers with 128 filters because that reduces the model input to a zero dimensional vector.}. Interestingly, bandwidth and duration can be predicted with as few as 30 packets and increasing the number of packets does not considerably improve the accuracy. For traffic classification task, there is a significant performance improvement from 30 to 60 packets. However, the traffic class prediction does not get more accurate by observing more packets.

\begin{figure}
	\centering
	\includegraphics[width=3.0in]{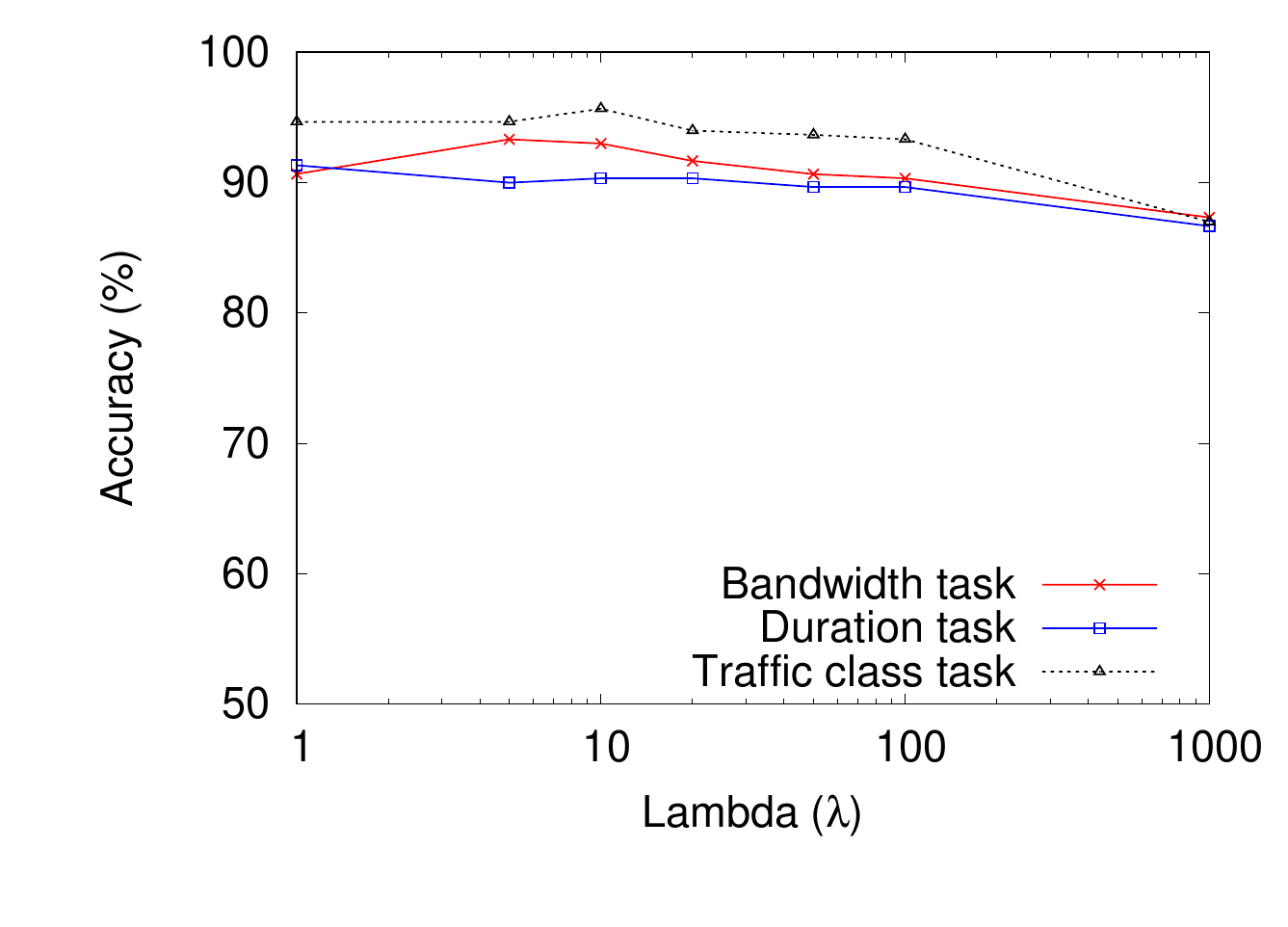}
	\caption{$\lambda$ versus accuracy for QUIC dataset}
	\label{fig-lambda-QUIC}
\end{figure}

Fig. \ref{fig-lambda-QUIC} shows the prediction accuracy of the three tasks with different $\lambda$. Intuitively, when the number of training samples of one task is considerably smaller than other tasks in multi-task learning, the shared parameters of the deep model are affected more by tasks with abundant data during training. Hence, increasing the weight of the loss function of the task with fewer data, as it is explained in Section \ref{model-arc}, may compensate for the lack of data during training and increase the effect of this task on the training procedure. As shown in Figure \ref{fig-lambda-QUIC}, increasing $\lambda$ helps the model to fit to the traffic class prediction tasks until it reaches the maximum accuracy. However, increasing the $\lambda$ further will degrade the accuracy of all tasks. That is because when $\lambda$ is very large, the model highly over-fits to the traffic classification training data and, consequently, performs worse on test data of all tasks. Moreover, when $\lambda$ is very large, the value of gradient updates becomes very large for traffic class prediction in comparison with other tasks which makes the training process extremely difficult to converge to local minimum without fine-tuning learning rate. This phenomenon affect the performance of all tasks. Hence, for the multi-task learning approach, one should find the suitable value of $\lambda$ as a hyper-parameter. A good starting point is to set $\lambda$ with the ratio of the number of samples of bandwidth and duration tasks over the number of samples of traffic classification task.


\begin{table*}[t]
	\centering
	\caption{Accuracy of QUIC dataset with different bandwidth and duration dividers}
	\label{tbl-QUIC-bw-duration}
	\begin{tabular}{*{15}{c}}
		Bandwidth divider (kbps) & Duration divider (s) & {Accuracy [Bandwidth, Duration, Traffic Class]}\\
		\hline
		[21.15, 164.02, 568.82, 2890.56] & [5.97, 21.35, 44.80, 84.00] & [90.67\%, 91.33\%, 94.67\%]  \\
		\hline
		[21.15, 164.02, 568.82, 2890.56] & [20, 40, 80, 100] & [91.33\%, 92.00\%, 93.33\%]  \\
		\hline
		[21.15, 164.02, 568.82, 2890.56] & [1, 50, 100, 150] & [91.33\%, 94.00\%, 92.66\%]  \\
		\hline
		[10, 50, 100, 1000] & [5.97, 21.35, 44.80, 84.00] & [92.00\%, 90.66\%, 94.00\%]  \\
		\hline
		[50, 100, 200, 300] & [5.97, 21.35, 44.80, 84.00] & [84.66\%, 90.00\%, 93.00\%]  \\
		\hline
		[50, 100, 200, 300] & [1, 50, 100, 150] & [82.00\%, 93.00\%, 92.00\%]  \\
		\hline
	\end{tabular}
\end{table*}

Table~\ref{tbl-QUIC-bw-duration} shows the effect of bandwidth and duration dividers on the performance of bandwidth, duration, and traffic class tasks. The first row shows the optimal dividers which are obtained from the training set as explained in Section \ref{sec-inp-out}. In this experiment, we only use 20 labeled samples from each class for training traffic class task. As shown, the optimal dividers achieve the highest accuracy on traffic class task. By changing the value of dividers, the accuracy of traffic class task slightly degrades. However, the accuracy of traffic class task is still considerably larger than single-task learning ($66.67\%$) and transfer learning ($87.33\%$), which are reported in Table~\ref{tbl-QUIC-accuracy}. In other word, even with non-optimal dividers, using multi-task learning is still beneficial and outperforms other approaches. 

This is useful because for some applications and scenarios, such as in ISPs and for resource allocation, the ISP may need to define a custom bandwidth classes to fulfill its need that may be different from the optimal divider obtained based on training data. Hence, not only the multi-task learning approach can be used to predict bandwidth and duration which can be useful for resource allocation or QoS provisioning, but it also improves the accuracy of the traffic class prediction regardless of the choice of dividers. In other words, the multi-task learning approach effectively kills two birds with one stone. Note that an arbitrary bad choice of dividers may lead to lower accuracy on the bandwidth and duration tasks, as in the last row of Table~\ref{tbl-QUIC-bw-duration}. But, it still improves the accuracy of traffic class because it helps the model to learn the shared parameters of the model with abundant data. Hence, if the bandwidth and duration tasks are planned to be used as auxiliary tasks to improve the accuracy of traffic class task, it is better to choose the optimal divider, as explained in Section \ref{sec-inp-out}. If the bandwidth and duration predictions are planned to be used for resource allocation or QoS proposes, defining the problem in a multi-task framework still improves the accuracy of traffic classification task.

\subsection{ISCX Dataset}
In this section, we use ISCX dataset and combine all classes into 5 different traffic types as explained in Section \ref{sec-datasets}. Although it has been shown that a CNN model with payload information as input achieves higher accuracy for this dataset \cite{lotfollahi2017deep}, we conduct experiments with ISCX dataset to only show the performance improvement of our multi-task learning approach over single-task learning and transfer learning approach in general. 

Table \ref{tbl-ISCX-accuracy} presents the accuracy of bandwidth, duration, and traffic class tasks for ISCX dataset. Regardless of the learning approach, the accuracy of traffic class task is lower in ISCX dataset (Table \ref{tbl-ISCX-accuracy}) than QUIC dataset (Table \ref{tbl-QUIC-accuracy}). Similar to Section \ref{sec-QUIC-eval}, we set the $\lambda$ to one and $k$ to 60 packets. As shown in Table \ref{tbl-ISCX-accuracy}, the accuracy of bandwidth and duration tasks are similar for multi-task learning and single-task learning approaches. This suggests that patterns and convolutional filters that are suitable for bandwidth prediction are also suitable for duration prediction because in multi-task learning approach they shared the same model parameters and they achieve the same accuracy. The accuracy of the traffic class task is not as high as other tasks, but the table shows a significant improvement with multi-task learning approach in comparison with other approaches.

\begin{table*}[t]
	\centering
	\caption{Accuracy on ISCX dataset}
	\label{tbl-ISCX-accuracy}
	\begin{tabular}{*{15}{c}}
		\hline
		- & \multicolumn{3}{c}{Accuracy [Bandwidth, Duration, Traffic Class]}\\
		\hline
		Number of labeled samples (For traffic class) & CNN+RNN-2 \cite{lopez2017network} & Transfer learning & Multi-task learning\\
		\hline
		10 & [88.33\%, 90.00\%, 52.33\%] & [-, -, 54.67\%] & [88.67\%, 90.00\%, 60.00\%]  \\
		20 & [88.33\%, 90.00\%, 57.00\%] & [-, -, 62.00\%] & [88.00\%, 89.33\%, 65.33\%]  \\
		50 & [88.33\%, 90.00\%, 60.67\%] & [-, -, 69.33\%] & [91.33\%, 90.00\%, 72.67\%]  \\
		100 & [88.33\%, 90.00\%, 77.33\%] & [-, -, 79.33\%] & [89.33\%, 91.33\%, 80.67\%]  \\
		\hline
	\end{tabular}
\end{table*}

\begin{figure}
	\centering
	\includegraphics[width=3.0in]{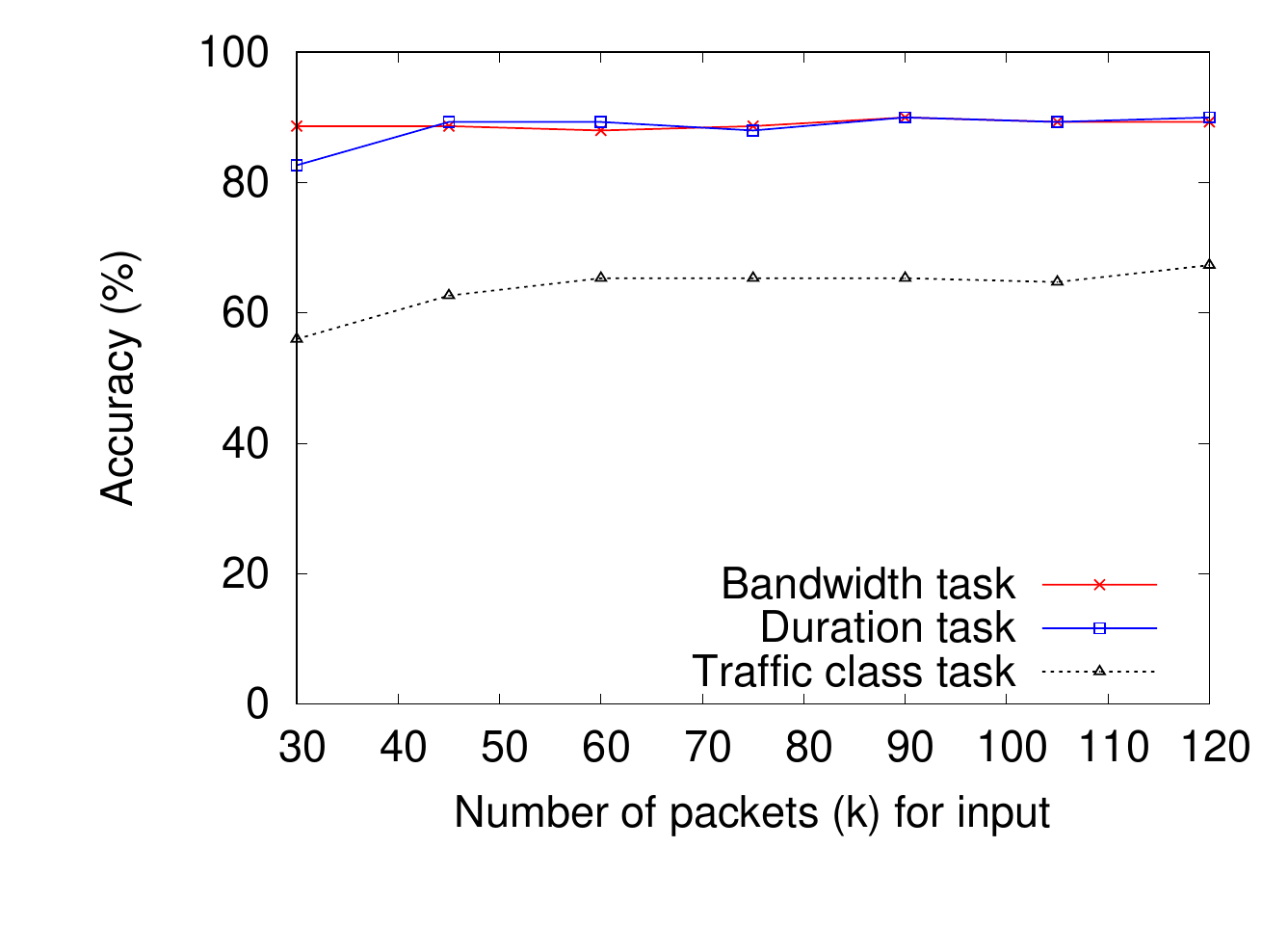}
	\caption{Number of input packets versus accuracy for ISCX dataset}
	\label{fig-acc-ISCX}
\end{figure}

Figure \ref{fig-acc-ISCX} illustrates the accuracy of the three tasks versus the number of packets, $k$. Unlike QUIC dataset, the duration task needs to observe more packets for accurate prediction. In QUIC dataset, the duration task accuracy is almost the same for different number of packets. However, For ISCX dataset, $k=30$ leads to much lower accuracy than $k>=40$. Bandwidth and traffic class prediction tasks show the same trend as QUIC dataset. Interestingly, both ISCX and QUIC datasets almost achieve their maximum accuracy for all their tasks with around 60 input packets.

\begin{figure}
	\centering
	\includegraphics[width=3.0in]{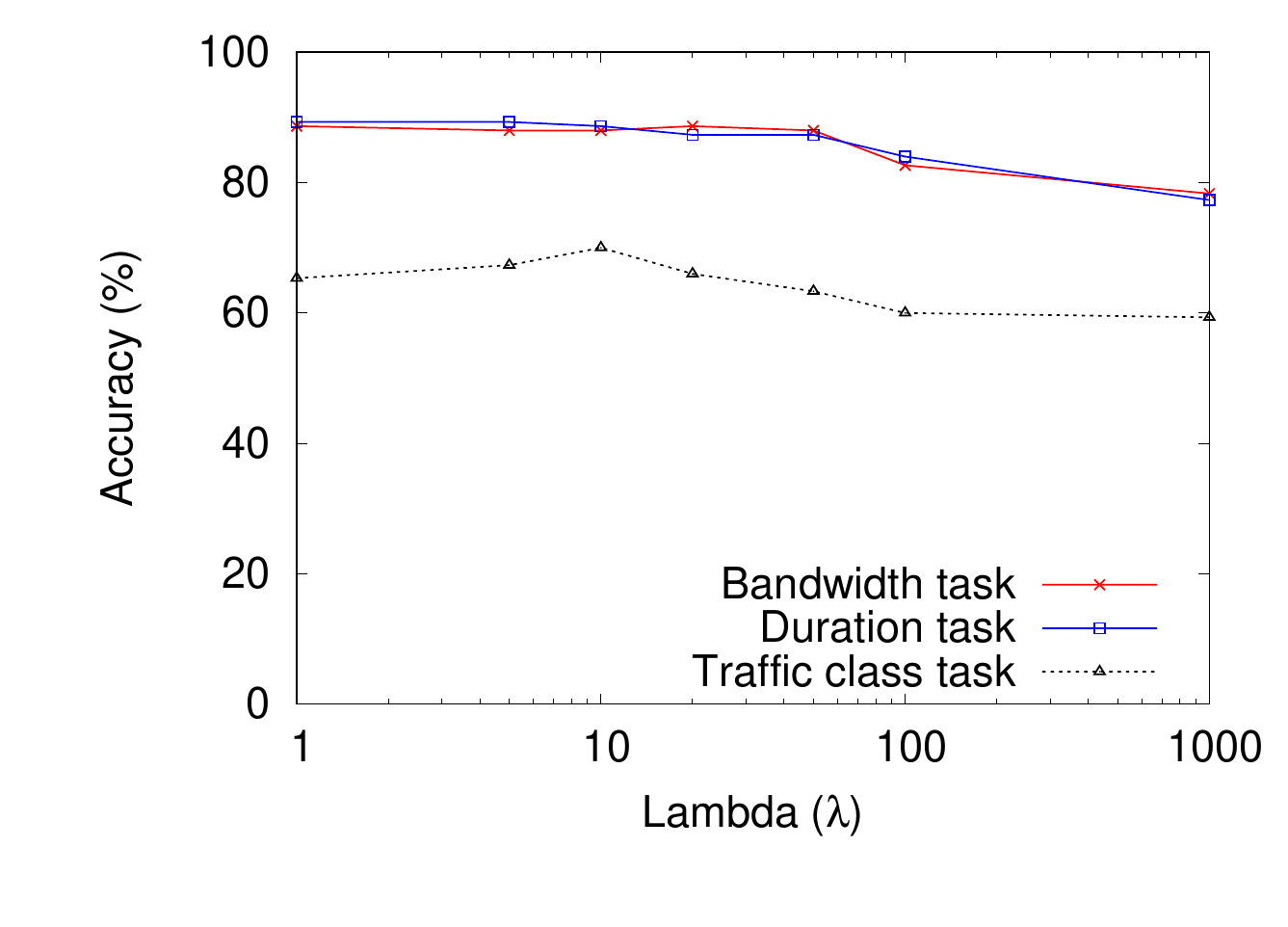}
	\caption{$\lambda$ versus accuracy for ISCX dataset}
	\label{fig-lambda-ISCX}
\end{figure}

Figure \ref{fig-lambda-ISCX} shows the accuracy of all tasks when using different $\lambda$. In this experiment, we use 20 labeled data samples per class (for traffic class prediction) and the entire dataset for bandwidth and duration tasks. Similar to QUIC dataset (Figure \ref{fig-lambda-QUIC}), the maximum accuracy of the traffic class prediction reaches around $\lambda = 10$. Similarly, as in Section \ref{sec-QUIC-eval}, by further increasing $\lambda$, the model over-fits to traffic classification task which degrades the performance of all tasks.

\section{Discussion}
In this paper, we use bandwidth and duration predictions as tasks with abundant training data. These two tasks have potential usage, as discussed earlier. However, for scenarios where these predictions are not important and they only serve to improve the accuracy of traffic classification, one can also use other prediction tasks (e.g. average inter-arrival time or number of bursts) as auxiliary tasks. Auxiliary tasks should have two characteristics: First, it should be highly relevant to the traffic classification task. Second, the label should be easy to obtain. In such cases, finding the best set of auxiliary tasks to improve the traffic class prediction should be treated similar to hyper-parameter tuning. The study and analysis of various auxiliary tasks are out of the scope of this paper and is considered as future work.

In our experiments, our multi-task learning approach outperforms, or performs as accurately as, the transfer learning approach. However, in many cases, transfer learning is the only option. For instance, if the pre-trained model is given without the training data, we can only use transfer learning. Additionally, if the time or computational complexity of training process is important, it may be desirable to avoid multi-task learning since it needs to train the whole model using both unlabeled data and labeled data. If a pre-trained model is available, transfer learning can train a model extremely fast, although using a public pre-trained model is shown to expose security threat \cite{rezaei2019target}. Otherwise, the multi-task learning framework is more effective than the transfer learning approach.

\section{Conclusion}
In this paper, we propose a multi-task learning approach that predicts traffic class labels as well as bandwidth and duration of network traffic flows. Because the bandwidth and duration tasks do not require human effort or controlled and isolated environment for labeling, a large amount of data can be easily captured and used for training these two tasks. We show that by providing a large enough dataset for bandwidth and duration tasks, one can train the traffic class prediction task with only a small number of samples. Hence, it obviates the need for a large amount of labeled data samples for traffic classification. Moreover, bandwidth and duration predictions can be used for resource allocation, routing and QoS purposes in ISPs. We conduct experiments with two public datasets: QUIC and ISCX VPN-nonVPN. We illustrate that our multi-task learning approach significantly outperforms both single-task and transfer learning approaches.

\section*{Acknowledgment}
This work was supported by the National Science Foundation (NSF) under Grant CNS-1547461, Grant CNS-1718901, and Grant IIS-1838207.

\end{document}